\documentclass[letterpaper]{sig-alternate-2013}

\permission{\copyright 2017 International World Wide Web Conference Committee (IW3C2), 
published under Creative Commons CC BY 4.0 License.}
\conferenceinfo{WWW 2017,}{April 3--7, 2017, Perth, Australia.}
\copyrightetc{ACM \the\acmcopyr}
\crdata{978-1-4503-4913-0/17/04. \\
http://dx.doi.org/10.1145/3041021.3054141 \\
\includegraphics{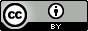}
}

\usepackage{graphicx}
\usepackage{multirow}
\usepackage{multicol}
\usepackage{array}
\usepackage{subfigure}

\usepackage{algorithm}
\usepackage{algorithmicx}
\usepackage{algpseudocode}
\usepackage{amsmath}
\usepackage{graphics}
\usepackage{epsfig}

\usepackage{xcolor,colortbl}
\usepackage{xcolor}
\usepackage{hhline}
\usepackage{makecell}

\begin{document}

\title{When Fashion Meets Big Data: Discriminative Mining of Best Selling Clothing Features}

\numberofauthors{2} 
\author{
\alignauthor
Kuan-Ting Chen\\
       \affaddr{National Taiwan University}\\
       \affaddr{Department of Computer Science}\\
       \affaddr{Taipei 10617, Taiwan}\\
       \email{ktchen@cmlab.csie.ntu.edu.tw}
\alignauthor
Jiebo Luo\\
       \affaddr{University of Rochester}\\
       \affaddr{Department of Computer Science}\\
       \affaddr{Rochester, New York 14627, USA}\\
       \email{jluo@cs.rochester.edu}
}

\maketitle
\begin{abstract}
With the prevalence of e-commence websites and the ease of online shopping, consumers are embracing huge amounts of various options in products. Undeniably, shopping is one of the most essential activities in our society and studying consumer's shopping behavior is important for the industry as well as sociology and psychology. Indisputable, one of the most popular e-commerce categories is clothing business. There arises the needs for analysis of popular and attractive clothing features which could further boost many emerging applications, such as clothing recommendation and advertising. In this work, we design a novel system that consists of three major components: 1) exploring and organizing a large-scale clothing dataset from a online shopping website, 2) pruning and extracting images of best-selling products in clothing item data and user transaction history, and 3) utilizing a machine learning based approach to discovering fine-grained clothing attributes as the representative and discriminative characteristics of popular clothing style elements. Through the experiments over a large-scale online clothing shopping dataset, we demonstrate the effectiveness of our proposed system, and obtain useful insights on clothing consumption trends and profitable clothing features.
\end{abstract}


\begin{figure}[!t]
\centering
\includegraphics[width=3.25in, height=1.7in]{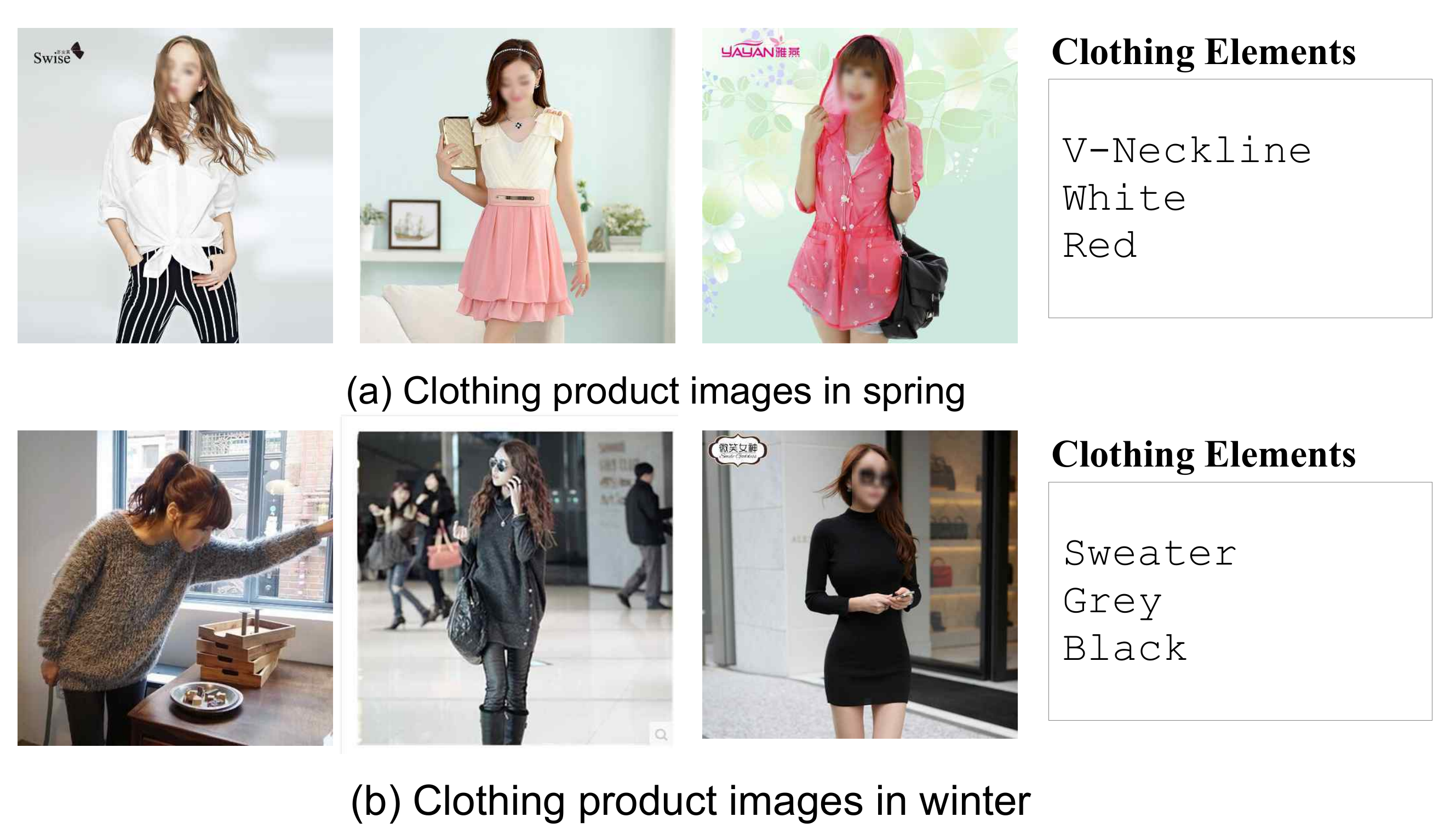}
\caption{Examples for the popular clothing items (a) in spring and (b) in winter on an online clothing shopping website.}
\label{fig:fig1}
\end{figure}

\section{Introduction}
Due to the boom of online shopping services, clothing business is one of the fastest-growing ventures in industry and technology today \cite{report:BBC}\cite{report:fashiontrend}\cite{report:NYCEDE}\cite{report:NYT}, as well as one of the most promising profitable platforms. The contributing factors for such tremendous growth include consumers' frequent adoption of broadband networks and mobile devices, changes in internet content, subsequent experiences on online shopping, and the constant upgrading of the online shopping process and convenience. eMarketers  reported that e-commerce sales will reach \$1.922 trillion in 2016 and increase nearly 23\% to \$2.356 trillion in 2018 \cite{report:eMarket}. Nielsen showed that the most popular e-commerce categories growing in prominence for online shopping including clothing, and airline and hotel reservations \cite{report:Nielsen}. Investigating effective of determination clothing selling items has become a great interest for the industry because of its promising opportunity for online shopping profit and for boosting many emerging applications such as clothing recommendation and advertising by clothing brand association. A traditional way to discover clothing selling trend and favorable style elements would be relying on the manual observation by experts or user survey. However, it is very time consuming and would vary with the season. 

In the academia field, there has been increasing interest in clothing product analysis from the computer vision and multimedia communities. The research closely related to our work could be mainly classified into two categories: clothing fashion analysis and product feature analysis by customer reviews.  For clothing fashion analysis, most existing fashion analysis works focused on the investigation of the clothing attributes, such as clothing parsing \cite{RW:BourdevMM11}\cite{RW:LiuCVPR12}\cite{RW:NguyenMM12}\cite{RW:TMM14}\cite{RW:liu2016deepfashion}, fashion trend \cite{hidayati:TrendNY} \cite{Chen:FashionPrada} and clothing retrieval \cite{RW:KiapourICCV2015}\cite{RW:LiuMM12}. For product feature analysis by customer reviews, the research studies \cite{RW:huAAAI04}\cite{RW:kumarIJWST12}\cite{RW:karkareicesc14}\cite{RW:humingMM04} considered customer reviews and proposed systems to summarize all the customer reviews of a product. However, the customer reviews might be noisy, ambiguous and inconsistent to a clothing producer  \cite{website:zappos}.

In contrast to other work, we focus on analyzing and learning the profitable clothing features by popular and attractive clothing features discovery (cf. Fig. \ref{fig:fig1}). Moreover, to our best knowledge, this is the first work to address the profitable clothing features in a major large-scale clothing shopping website. In this paper, we first organize a large-scale Alibaba Taobao Clothing Dataset: a large number of clothing data with customers' transaction history from a real-world large-scale online shopping website, Taobao. We then exploit and analyze attractive and profitable clothing features in this large-scale clothing dataset. Moreover, the clothing features are extracted by automatically analyzing clothing images. More specifically, for every image, we automatically extract 60 clothing attributes such as collar, necktie, color, etc. Using semantic clothing attributes to represent clothing products can tell online sellers the most popular clothing elements, which is not only a specific clothing reference to understand customers' preference but also is a good way to consider clothing elements for clothing designers in the view of industry. In our experimental results, we demonstrate the effectiveness of the clothing attributes and further analyze the profitable clothing features. 

We presented the preliminary results in \cite{Chen:FashionPrada}. In this paper, we propose to 1) prune noisy images by a deep learning approach (cf. Section \ref{subsec:approach_prune}), 2) incorporate the product sales information from an online business platform to measure the true impact of clothing elements on consumers (cf. Section \ref{subsec:approach_Mining}), 3) investigate the effects of clothing attribute representation on a large-scale online shopping dataset (cf. Section \ref{sec:dataset}), and 4) provide more details on the proposed approaches (cf. Section \ref{subsec:approach_featlearn}), experimental results, and discussions (cf. Section \ref{sec:results}).

The primary contributions of this paper include:

\begin{itemize}
\item Proposing a framework that facilitates the investigation of consumers' clothing preference in a fine-grained manner (Section \ref{sec:framework}). 
\item Conducting empirical analysis of a large-scale online shopping dataset collected between June 2014 and June 2015 (Section \ref{sec:dataset}).
\item Implementing an effective and efficient method for pruning noisy images in the online shopping dataset (Section \ref{sec:approach}).
\item Mining attractive and profitable clothing features in a large number of clothing data with customers' transaction history (Section \ref{sec:approach}). 
\item Discovering significant insights using the proposed framework from real-world large-scale data (Section \ref{sec:results}).
\end{itemize}

\begin{figure}[!t]
\centering
\includegraphics[width=3.3in]{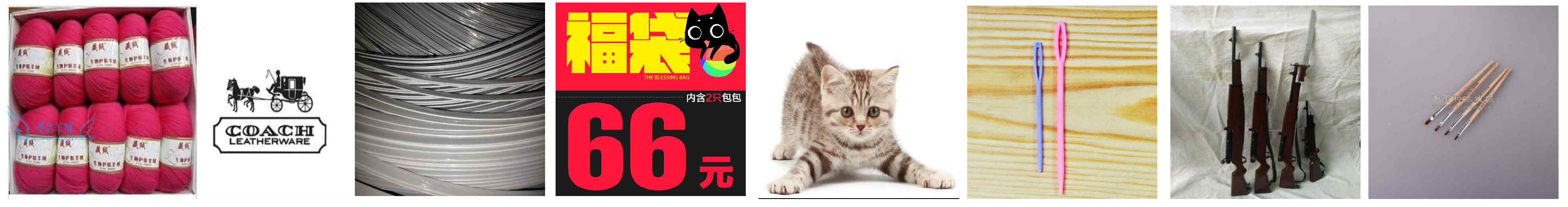}
\caption{Examples of noisy images in an online clothing dataset. These noisy images are not clothing items but might be shown along with an popular clothing item for a clothing product, such as an advertisement.}
\label{fig:noise}
\end{figure}

\begin{figure*}[!t]
\centering
\includegraphics[width=7in, height=3in]{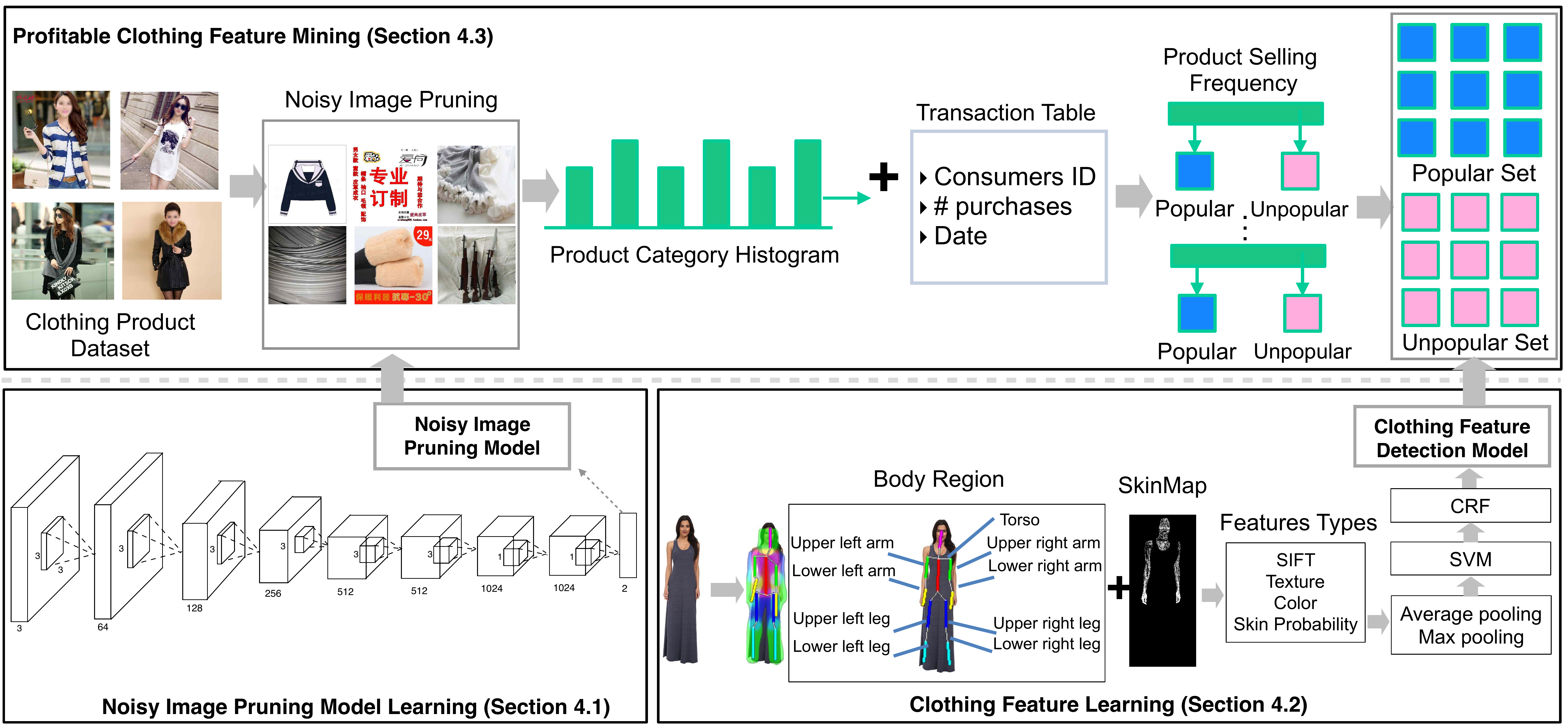}%
\caption{An illustration of our proposed framework. Our system includes three major components. Top: Profitable clothing features mining, Bottom Left: Noisy image pruning model learning. Bottom Right: Clothing feature learning. The shopping transactions analysis can specify the selling frequency of each product item to further construct popular and unpopular product item sets. The Noisy image pruning could clean noisy, unrelated, and inappropriate clothing images and the clothing feature learning will conduct mining informative clothing features.}
\label{fig:flow}
\end{figure*} 

\section{Overview of the framework}
\label{sec:framework}
To discover popular and attractive clothing features, an effective framework of analyzing clothing shopping transactions to draw profitable clothing features from a large-scale dataset is beneficial for an online shopping industry. The proposed system diagram is shown in Fig. \ref{fig:flow}. The core algorithms include:

(a) Noisy image pruning model learning. We observed that a part of images are not clothing items but might be shown along with an attractive clothing item for a clothing product, such as an advertisement, pets, and the logo of a clothing brand. These noisy, unrelated, and inappropriate images are referred to as \textit{noisy images} in our work (cf. Fig. \ref{fig:noise}). To tackle this problem, the intuitive way is browsing the whole dataset and manually filtering noisy images. However, this is very time consuming for a large-scale dataset and restricts the scalability of the system. The pruning of noisy images can be treated as a binary classification problem. Inspired by the deep learning architecture, which has achieved very promising results in handwritten digits \cite{RW:lecun1998gradient}, image classification \cite{RW:ILSVRC15}, speech recognition \cite{RW:dahl2010phone}, computer vision \cite{RW:krizhevsky2012imagenet} and natural language processing \cite{RW:collobert2008unified}, we learn a classifier based on a deep learning architecture to automatically pruning noisy images (cf. Section \ref{subsec:approach_prune}). 

(b) Clothing feature learning. Using an appropriate clothing representation for exploring clothing style characteristic is required to offer a semantic and intuitive way to determine what clothing elements people would like to purchase. Inspired by the paper \cite{Chen:clothAttri}, a learning-based clothing attributes approach was carried out to describe clothing style. In the research \cite{Chen:clothAttri}, Chen et al. only detected 42 upper body clothing attributes. We observed the clothing information in the lower body is an essential clue for clothing style understanding (e.g. pants or skirt). Motivated by the research \cite{hidayati:TrendNY}, we utilize New York Fashion Show images for learning clothing style features, which contains 3914 images from 2014 summer/spring New York Fashion Show and 4000 images from 2015 summer/spring New York Fashion Show, respectively \cite{dataset:vogue}. The 7914 images from 2014 and 2015 New York Fashion Shows are used to extract features and learning 60-attribute semantic representation to describe both upper and lower body clothing features (cf. Section \ref{subsec:approach_featlearn}). 

(c) Profitable clothing feature mining. First, we eliminate noisy and unrelated clothing images on the online shopping dataset using noisy image pruning model. Then, we split clothing items into different bins of a category histogram. In order to measure the popularity of clothing items, we then extract the selling frequency of each clothing item from user transaction history table. Next, we exploit and analyze the popularity of clothing items in different seasons, followed by clothing feature extraction as the representation of clothing style features  (cf. Section \ref{subsec:approach_Mining}). In the following, we first describe clothing datasets and then the adopted approaches for mining the profitable clothing features.

\begin{table*}[!t]
\renewcommand{\arraystretch}{1.05}
\caption{A summary of clothing item categories.}
\label{tab:category}
\centering
\begin{tabular}{|p{1cm}<{\centering}|p{1.7cm}<{\centering}|p{2cm}<{\centering}|p{1.9cm}<{\centering}|p{2.2cm}<{\centering}|p{1.1cm}<{\centering}|p{1.3cm}<{\centering}|p{1.5cm}<{\centering}|p{1cm}<{\centering}|}
\hline
\multirow{2}[4]{1cm}{\textbf{Upper body}}
    & Coat & T-shirt & Shirt & Spaghette  & Smock & Tank & Sweater & Collar \\ 
\cline{2-9} 
    & Underwear & Sport & Winter & Raincoat & Leather & Suit & Trench & Furs\\
\hline
\multirow{2}[4]{1cm}{\textbf{Lower body}}
    & Pants & Legging & Skirt & Bloomers & Wedding & Jeans & Briefs & Silk \\
\cline{2-9}    
    & Short & Casual Shoes &  Rainy Shoes & Sports Shoes & Boots  & Slipper & & \\
\hline
\multirow{2}[6]{1cm}{\textbf{Whole body}}
    & Suit & Pajamas & Sport & Sun Protection & Uniform  &  Wedding   &  Chenogsum &  Dress  \\ 
\cline{2-9}    
    & Work (Server) &  Work (Doctor) &   Activewear (Cheer) &  Activewear (Performance)& & & & \\

\hline
\end{tabular}
\end{table*}

\section{Clothing DATASET Collection}
\label{sec:dataset}
In this work, we conduct our experiments on two datasets.

\begin{figure}[!t]
\centering
\includegraphics[width=3.3in, height=1.6in]{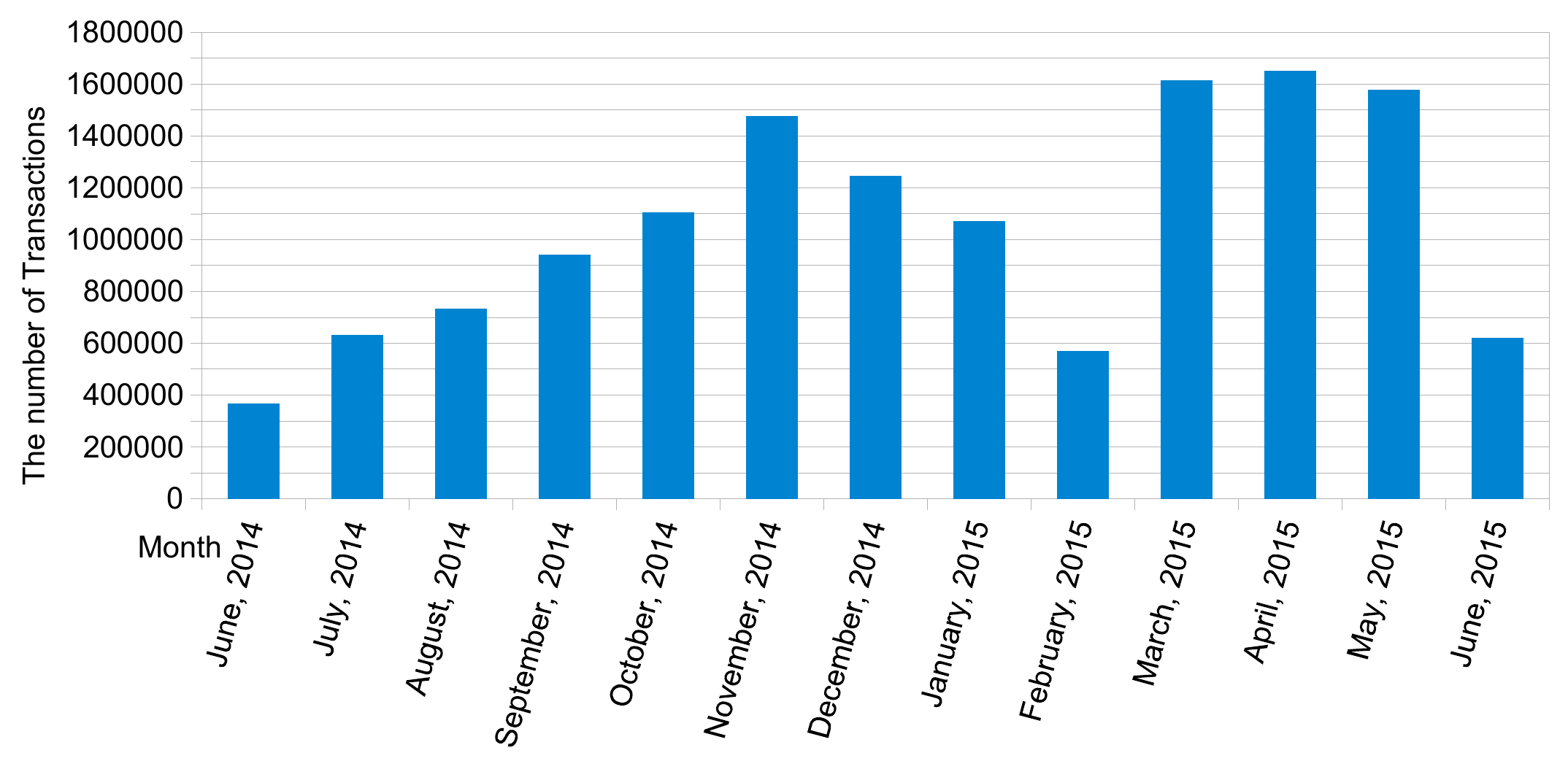}
\caption{The number of transactions in each month from June 2014 to June 2015 in Taobao clothing shopping dataset. The transaction information could reflect the real impact of a clothing item on customers in different seasons.}
\label{fig:numtrans}
\end{figure}

\textbf{1. Online Clothing Shopping Dataset.} In order to study the feasible and popular clothing features, we mainly exploit the profitable clothing features in a large-scale clothing shopping platform. Taobao is one of the largest online shopping website in China, which is similar to eBay and Amazon. In 2015, Taobao released a large-scale clothing dataset which includes clothing collocation from fashion experts, image data of Taobao items, and user behavior data. The item data table, item image, and user transaction history are utilized in this work. Examples of clothing product images are shown in Fig. \ref{fig:fig1} and Fig. \ref{fig:flow}.  In particular, \textbf{ 1) the item data table} contains about half million clothing products sold on Taobao during 13 months from  June 2014 to June 2015. In this table, there are four types of data: item\_id is a unique id for each product, cat\_id is the category id the product belongs to, name\_arr is an array that contains the name of this product and img\_data is the image information of each product. Note that we observe that the category id in this table is only a number and a large number of irrelevant clothing items fall into the same category. Therefore, we define new clothing categories with more semantic meaning in this work (cf. Table \ref{tab:category}). \textbf{(2) The item image} contains images for each product in item data table. Some images only present a single item and some images have models that wear the items in order to show the try-on style. \textbf{(3) The user history table} contains around 10 millions user transaction data. In this table, there are three types of data: user\_id is the user's unique id in one transaction, item\_id is the id of specific product the user purchases in this transaction, and the date is the time information of this transaction. The number of transactions in each month is shown in Fig. \ref{fig:numtrans}.

\textbf{2. New York Fashion Show.} In order to learn clothing feature detection models, we manually labeled the entire clothing dataset \cite{Chen:FashionPrada} for complete fine-grained clothing attribute annotations. This dataset contains 3914 images from 2014 summer/spring New York Fashion Show and 4000 images from 2015 summer/spring New York Fashion Show, respectively \cite{dataset:vogue}. The 7914 images from 2014 and 2015 New York Fashion Shows are used to extract features and to learn clothing attributes. 

\section{Discriminative mining of best \\ selling clothing features}
\label{sec:approach}

\subsection{Noisy Image Pruning}
\label{subsec:approach_prune}
As illustrated in Fig. \ref{fig:flow}, we propose to prune noisy images from the clothing shopping dataset. This is based on the observation that these images are not clothing items but might be shown along with an attractive clothing item for a clothing product, such as an advertisement and the logo of a clothing brand (cf. Fig. \ref{fig:noise}). The intuitive way to tackle this problem is browsing the whole dataset and manually filtering noisy images . However, this is very time consuming for a large-scale dataset. Taking the scalability and generalization of the proposed system into consideration, we learn a classifier for automatically filtering noisy images. 

The deep learning framework is considered as one promising direction by the research community and has been proven to be effective in various classification tasks, including handwritten digits \cite{RW:lecun1998gradient}, image classification \cite{RW:ILSVRC15}, speech recognition \cite{RW:dahl2010phone}, computer vision \cite{RW:krizhevsky2012imagenet} and natural language processing \cite{RW:collobert2008unified}. The noisy image pruning can also be treated as a binary classification problem.

The network structure we employ is similar to VGG-16 \cite{RW:simonyan2014very}, which has been demonstrated powerful in various computer vision tasks. First, we resized each image to 256$\times$256 and the resized images are processed by five convolutional layers. Each convolutional layer is also followed by max-pooling layers. Max-pooling is performed over a 2$\times$2 pixel window, with a stride of 2 pixels. A stack of 5 convolutional layers is followed by three fully connected layers. The first two layers have 4096 kernels each and are followed by dropout regularizations \cite{RW:srivastava2014dropout}. The final fully connected layer performs a softmax activation with 2 kernels for the neurons to turn real-valued vector into a vector of probabilities. We use rectified linearunits (ReLUs) activation functions \cite{RW:nair2010rectified} for first 5 convolution layers and 2 fully connneted layers. Furthermore, we utilize the cross-entropy loss function during training, the preferred loss function for binary classification problems. The model also uses the efficient Adam optimization algorithm for gradient descent. This model for noisy image pruning is trained and implement using the Tensorflow \cite{RW:abadi2016tensorflow} backend with the batch size of 128. We manually labeled noisy images and randomly split them into 80\% training, 10\% validation, and 10\% testing. Our noisy image pruning model achieves an accuracy of 75.5\%, a recall of 70\%, and a precision of 78.6\%.

\subsection{Clothing Feature Learning}
\label{subsec:approach_featlearn}

\subsubsection{Pose Estimation and Body Region Extraction}
\label{subsubsec:pose_est}

In order to learn clothing features, we need to extract visual features beforehand to train classifiers for every clothing attributes. Thanks for Marcin Eichner's team \cite{Eichner:pose}, we apply their pose estimation software to detect the pose of a human body and retrieve the body region of the model. We briefly describe the method of the pose estimation. First, a human upper-body is detected by a pre-trained upper-body detector. More clearly, the approximate location and scale of the person, and where the torso and head should lie could be roughly determined by using a sliding window detection based on Histograms of Oriented Gradients. Next, the structure of the detection window is utilized as the initialization of a Grabcut segmentation \cite{RW:rother2004grabcut}. A human body could be represented as a pictorial structure composed of body parts tied together in a tree-structured. Therefore, given an image $I$ , the location and orientation of each body part $l_i$ could be inferenced by the posterior of a configuration of human body parts $L =\{l_i\}$ using a log-linear model:

\begin{equation} 
 P(L|I) \propto exp\bigg( \sum_{(i,j) \in E} \Psi (l_i,l_j) + \sum_{i} \Phi(l_i)\bigg)\bigg),
\end{equation} 

where the binary potential $\Psi (I_i,I_j)$ corresponds to a spatial prior on the relative position of parts, e.g., the upper arms must be attached to the torso, and the unary potential $\Phi(l_i)$ corresponds to the likelihood of a local image evidence for a part in a particular position. More specifically, the pose estimation process is to segment body into nine parts: torso, upper left arm, upper right arm, lower left arm, lower right arm, upper left leg, upper right leg, lower left leg and lower right leg. Furthermore, four different kinds of visual features are computed in each body part, including color in the LAB space, texture descriptors, SIFT local feature, and skin probabilities. Finally, the features are aggregated by employing average or max pooling to generate a visual feature vector for all the parts of the body. An example is illustrated in Fig. \ref{fig:flow}.

\subsubsection{Clothing Attribute Learning}
\label{subsubsec:approach_feature}
The most intuitive way for training each clothing attribute is concatenating 72 features (i.e. 9 human body parts, 4 different kinds of visual features and 2 aggregation methods) into a long vector that becomes the full body visual feature vector. However, the influence of different types of visual feature on each attributes may vary. For example, texture features might have a great effect on pattern based attributes. Consequently, we compute the classification performance of each feature to represent the importance of features as weights towards individual attributes. As a result, we adopt a Support Vector Machine (SVM) \cite{chang:svm} with a Chi-square kernel to learning 60 clothing attribute models \cite{Chen:FashionPrada} and a weighting parameter are applied to vectors from different features to emphasize the importance of different visual features.

\begin{table*}[!t]
\renewcommand{\arraystretch}{1.05}
\caption{The classic/attractive, popular and unpopular clothing features in spring and winter.}
\label{tab:popular}
\centering
\begin{tabular}{|c|c|c||c|c|c|}
\hline
\multicolumn{6}{|c|}{\textbf{Classic / Attractive}} \\
\hline
\multicolumn{6}{|c|}{white, black, multicolor, lower\_solid, round\_neckline}\\
\hline
\multicolumn{3}{|c||}{\textbf{Spring Popular}} & \multicolumn{3}{c|}{\textbf{Winter Popular}} \\ 

\hline
bags\_accessories & belt &  neckring                              & bags\_accessories & collar & placket\\
\hline
upper\_floral & upper\_graphics &  lower\_floral                 & upper\_graphics & lower\_graphics & upper\_gray\\
\hline
upper\_graphics & upper\_blue & upper\_red                        & upper\_yellow & lower\_gray & lower\_brown\\
\hline
v\_neckline  &  Other\_Neckline  &                                & other\_neckline& & \\

\hline
\multicolumn{3}{|c||}{\textbf{Spring Unpopular}} & \multicolumn{3}{c|}{\textbf{Winter Unpopular}} \\
\hline
 Collar & upper\_solid & lower\_brown                            & belt & upper\_solid & upper\_blue \\
\hline
lower\_gray  & &                                                  & upper\_red & lower\_blue & lower\_red\\
\hline
\end{tabular}
\end{table*}

\subsubsection{Attribute Relation Inference}
\label{subsubsec:approach_Infer}{}
In section \ref{subsubsec:approach_feature} the clothing features are considered as isolated attributes. However, it is highly possible that some attributes appear in pairs (or groups). For example, we observe that a plaid shirt might have more than two colors. Note that inter-attribute dependencies are not always symmetric. For example, while a plaid shirt strongly suggests the presence of more than two colors, more than two colors do not necessarily suggest a shirt being plaid. We adopt a Conditional Random Fields (CRF) approach to inference the relation betweens attributes. More specifically, each clothing attribute acts as a node in the CRF framework and the edge connecting every two nodes indicates the joint probability of these two attributes. We build a fully connected CRF with all the attributes pairwise connected. The conditional probability of two clothing attributes $\{A_i, A_j\}$ given features $\{f_a, f_b\}$ is maximized by:

\begin{equation} 
 P(A_i, A_j|f_a,f_b) \propto \frac{P(A_i|f_a)}{P(A_i)}\frac{P(A_j|f_b)}{P(A_j)} P(A_i, A_j).
\end{equation}

\subsection{Profitable Clothing Feature Mining}
\label{subsec:approach_Mining}
To exploit attractive and profitable clothing features, we extract and analyze popular clothing features in a large-scale online clothing shopping dataset. First, we split clothing items into groups using category information. More specifically, the clothing items are separated into different category bins. Table \ref{tab:category} shows the details of clothing product categories. In order to measure the popularity of clothing items, we then extract the selling frequency of each clothing item from user transaction history table. Moreover, we integrate the popularity information into category bins. In addition, the popularity of clothing item style and transaction might be various in different seasons. Therefore, we further take this context (e.g. Spring and Winter) into consideration in our system. Next, all clothing items are sorted based on selling frequency in each category bin and the major proportion (i.e. Top 10\% in our work) of clothing selling items in each bin are picked in different seasons as the popular clothing items. The clothing features are extracted from popular clothing items as profitable attribute references (cf. Section \ref{subsec:approach_featlearn}), which could be utilized not only to maximize the online clothing shopping system revenues, but also to provide sellers and designers a popular clothing style reference. Note that the price information might be one of influence factors for popularity. Due to lacking of price information in this dataset, we tentatively consider the selling frequency in this work but could flexibly combine price information in this framework. Finally, we adopt the Fg-growth algorithm \cite{RW:hanSIGMOD00} to extract the frequent item sets of clothing features to further discuss and analyze popular clothing features in different seasons.

\section{EXPERIMENTAL RESULTS}
\label{sec:results}
We conduct several experiments to gain understanding of the performance of the clothing feature detection models. The overall accuracy of models is 62.6\%. An interesting observation is that some categories suffer from worse results (e.g. 42\% accuracy of belt, 56\% accuracy of accessories) since the objects are relatively small compared to the entire body. In the future, we could segment the body into more parts to improve the accuracy of the feature detection model. For example, we could segment the middle body for belt and the neck part for the accessories. We divide these models into three categories, color, pattern and clothing style. We observed and discovered that these results can be a very significant reflection for the fashion shows' styles. For example, in upper body color, both 2014 and 2015 fashion images have a large amount of white, gray and black colors. In upper body and lower body patterns, solid pattern is the classic pattern and solid pattern clothes always dominate every year's fashion shows. In the style category, there are many images with skirt in spring/summer fashion shows. These examples indicate the clothing feature detection models are very reasonable and effective for clothing features representation. It is worth noting that we summarize the experimental results of the clothing detection model in this paper and the detailed discussion of quantitative results is in \cite{Chen:FashionPrada}.

\begin{figure*}[!t]
\centering
\includegraphics[width=3.45in, height=1.5in]{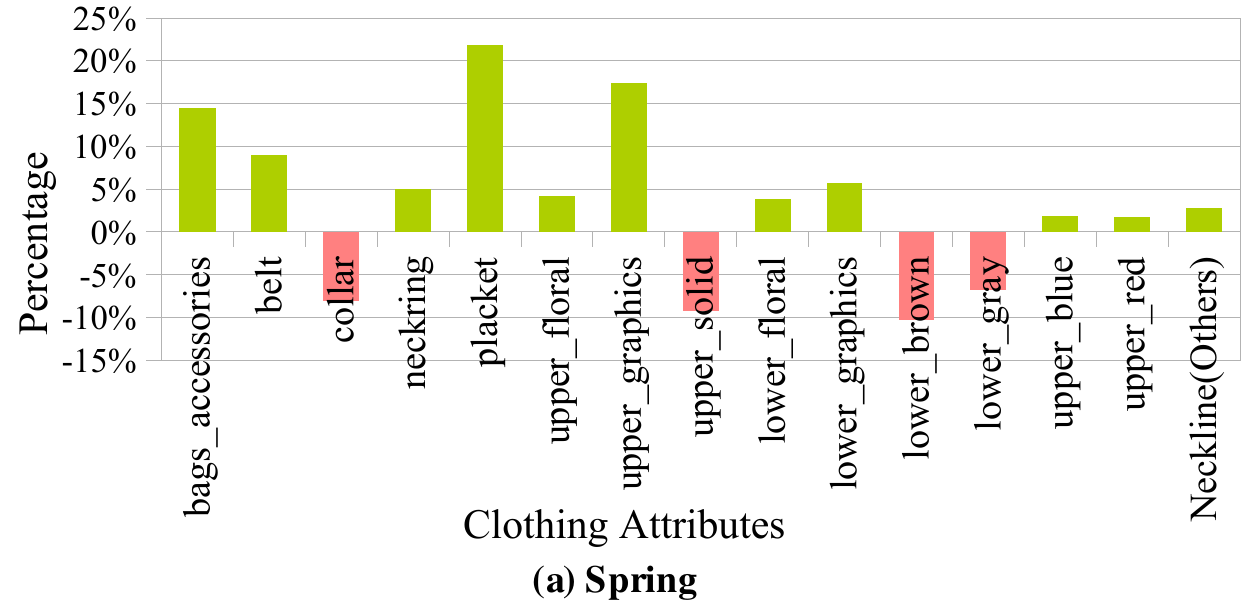}
\includegraphics[width=3.45in, height=1.5in]{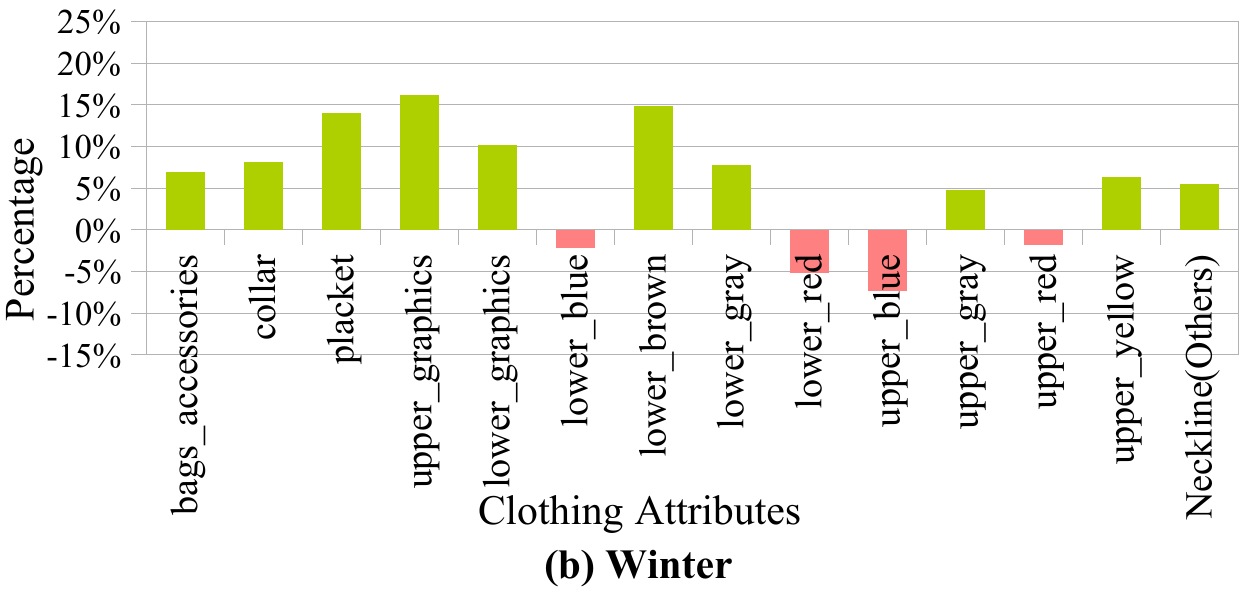}
\caption{The comparison of popular and unpopular clothing features (a) in spring and (b) in winter. The clothing features which boost customer's shopping behavior are marked green. The clothing features which lower customer's shopping behavior are marked red. Note that we remove the clothing features which has a little effect on customer's shopping behavior due to the limitation of paper length. More details and examples are presented online.\protect\footnotemark} 
\label{fig:eval_comb}
\end{figure*}

\begin{figure*}[!t]
\centering
\includegraphics[width=6.9in]{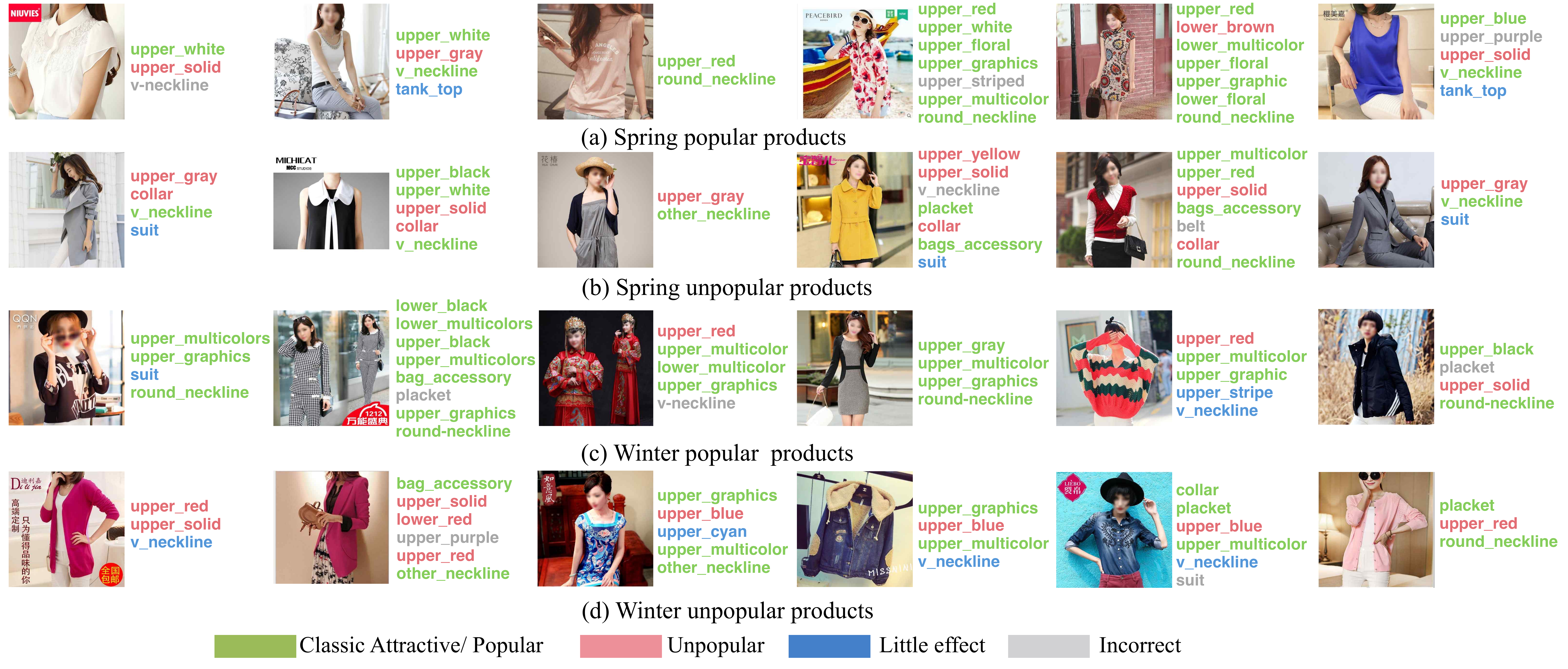}\\
\caption{Examples of mining clothing features showing (a) in spring popular products, (b) in spring unpopular products, (c) in winter popular products, and (d) in winter unpopular products.}
\label{fig:eval_ex}
\end{figure*}

Table \ref{tab:popular} shows classic/attractive, popular, and unpopular clothing features on the online clothing shopping dataset. We observed that there are some colors or styles that have large amount of image product items both in the frequent and seldom selling clothing items in 2014 winter and 2015 spring. For example, white, black, multicolor in both upper and lower body, solid pattern in lower body, and round-shape neckline. The white and black colors are reasonable because these two colors could be for all-purpose and be easily matched with other colors. Therefore, these two clothing features are most likely to appear during the whole year and referred to as \textit{classic clothing features} in our experiments. In addition, multicolor, lower\_solid, and round\_neckline presented in large numbers both in popular and unpopular clothing items in 2014 winter and 2015 spring as well. These clothing features could be regarded as attractive and safe features in the selling products, which are referred to as \textit{attractive clothing features} in our experiments.

Fig. \ref{fig:eval_comb} (a) compares popular clothing features with unpopular clothing features in spring. Note that we remove classic clothing features, black and white, and attractive clothing features, multicolor, lower\_solid, round\_neckline to present the changes more clearly. Interesting, people tend to wear blue and red colors in the upper body and are less likely to wear gray and brown colors in spring. A reasonable explanation might be that light-colored clothes can absorb less sun light instead of deeper color clothes, and therefore these clothes can keep people cooler in spring. Besides, graphics and floral patterns are popular clothing features. We observed that spring is a time of renewal and refresh; therefore these lovely patterns could be in line with the spring theme (cf. Fig. \ref{fig:eval_ex} (a)). These clothing features, which could boost customer's shopping behavior, are referred to as \textit{popular clothing features}. The visualizations of popular clothing features are shown in Fig. \ref{fig:vis_feature}. The clothing features such as brown and gray colors, which had led to a decrease in customer's consumption, are referred to as \textit{unpopular clothing features} (cf. Fig. \ref{fig:eval_ex} (b)).

\footnotetext{http://cmlab.csie.ntu.edu.tw/$\scriptsize{\sim}$ktchen/fashion.html}

\begin{figure*}[!t]
\centering
\includegraphics[width=6.7in, height=1.85in]{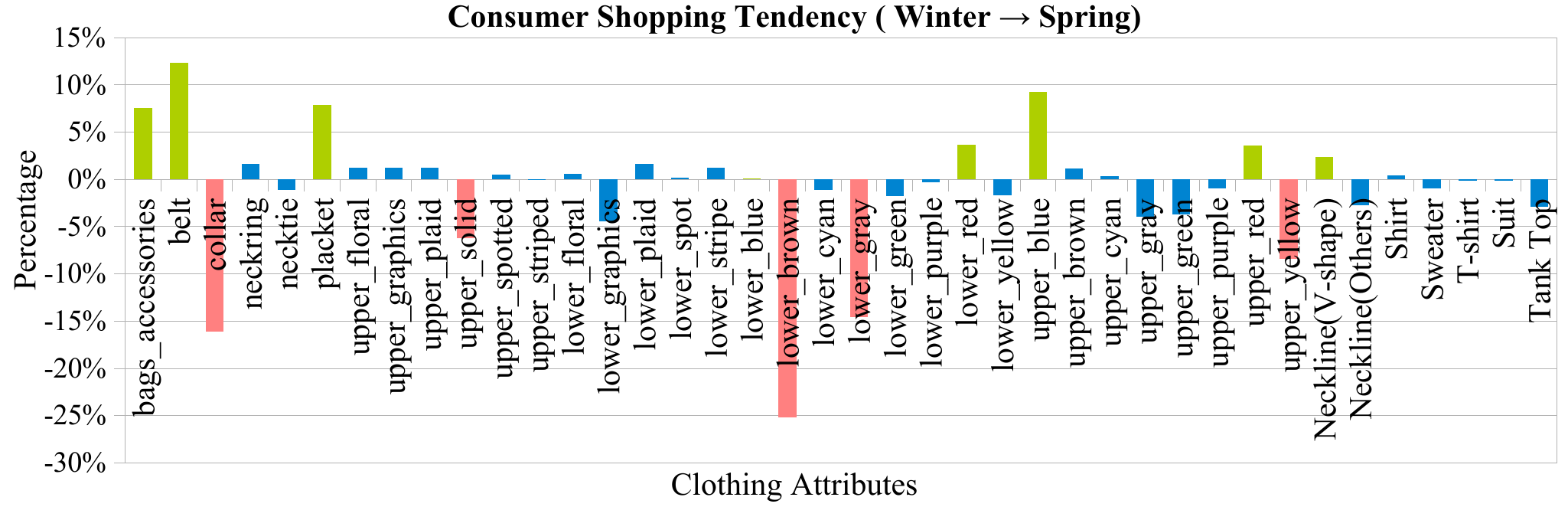}%
\caption{The comparison of clothing features change between spring and winter. The clothing features which appeared more frequently in spring are marked green. The clothing features the trend of which decreased in spring are marked red. The clothing features which had few differences between spring and winter are marked blue.}
\label{fig:eval_trend}
\end{figure*}

\begin{figure}[!t]
\centering
\includegraphics[width=3.2in]{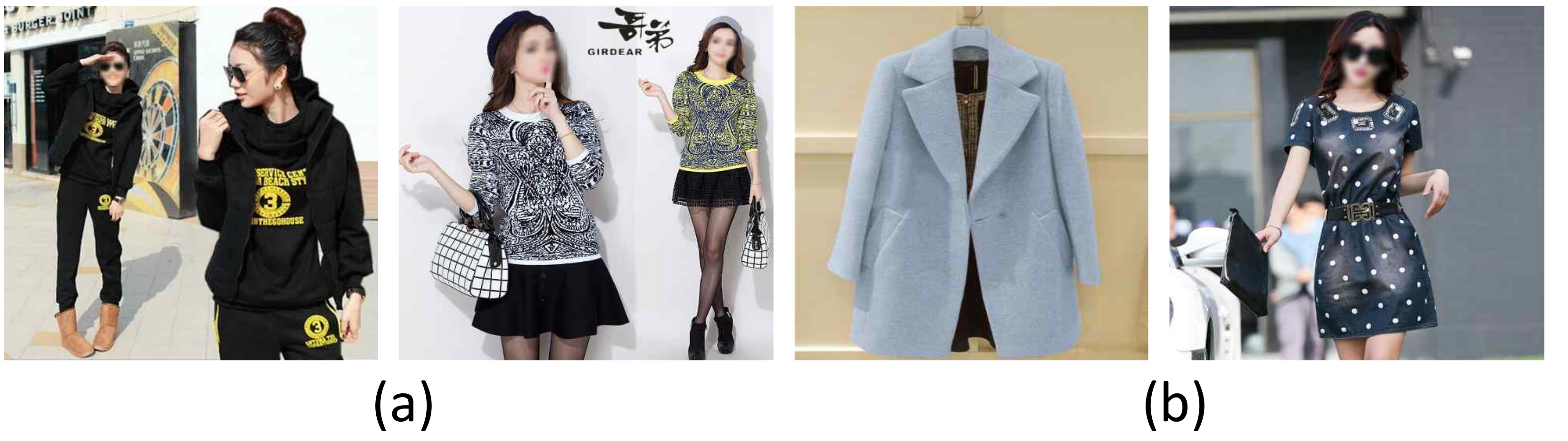}%
\caption{Examples of useful insights discovered by our proposed framework. (a)Unique clothing styles at the time. (b)Particular clothing outfits.}
\label{fig:eval_wrong}
\end{figure}

\begin{figure}[!t]
\centering
\includegraphics[width=2.4in]{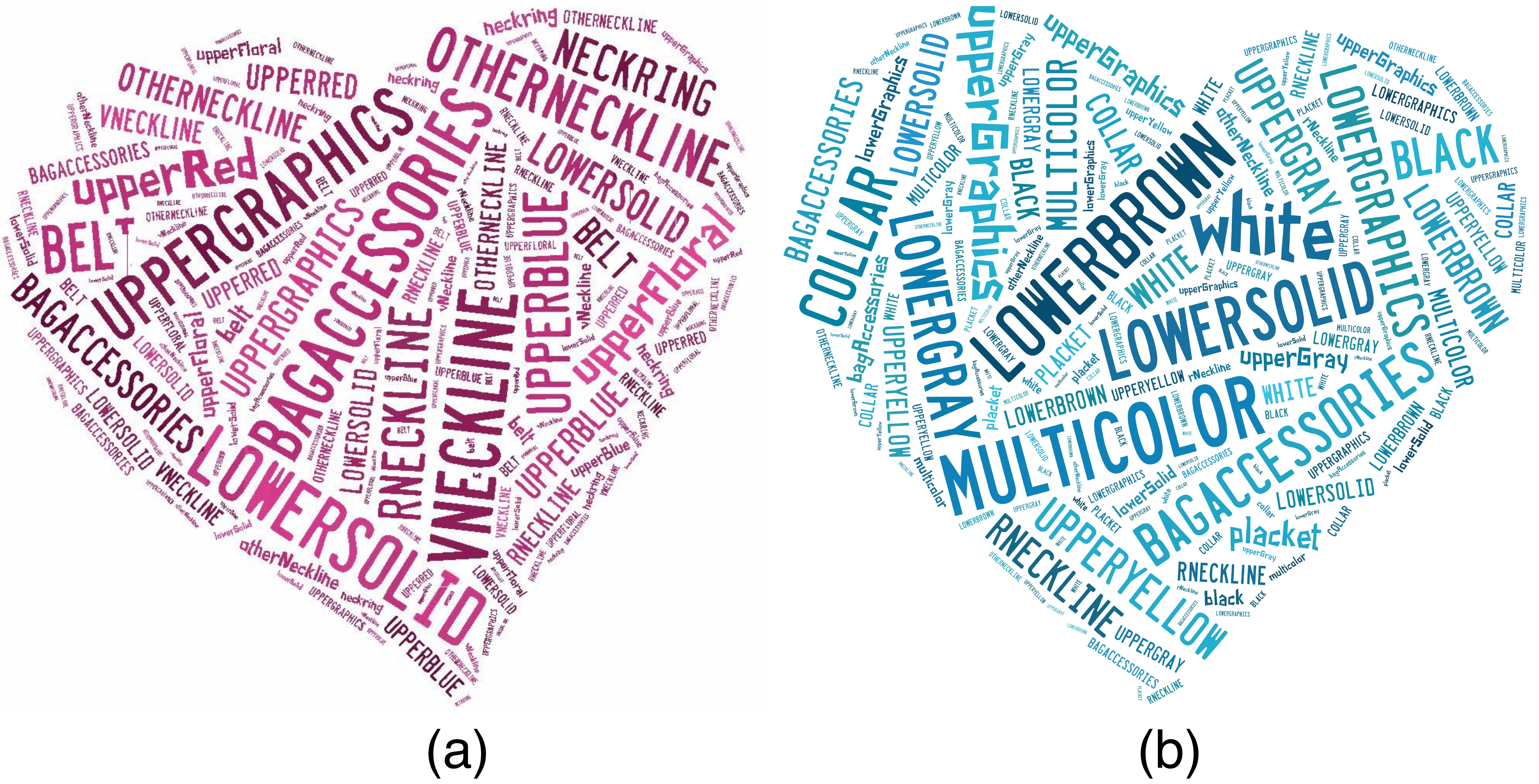}%
\caption{Visualization of the frequency of popular clothing features (a) in spring and (b) in winter. The fine-grained clothing representation in our framework could be a concrete reference to help sellers and designers understand customers' preference.}
\label{fig:vis_feature}
\end{figure}

Fig. \ref{fig:eval_comb} (b) compares popular with unpopular clothing features in winter. In contract to spring, the darker colors (e.g. brown and gray colors) and clothing styles which could keep people warmer (e.g. collar) are more popular. This phenomenon is typical to a colder season in a year. Interestingly, the yellow color could encourage more consumptions in winter. We observed that the yellow color is a small portion compared to the main color in an entire clothing product. This combination of clothing features (e.g. mainly black or darker colors mixed with a small portion of a light color as shown in Fig. \ref{fig:eval_wrong} (a) or an unpopular red color mixed with large portions of different colors in Fig. \ref{fig:eval_ex} (c)) could be regarded as unique clothing styles at the time and could be further provided to sellers for references to import proper clothing products.  Another interesting observation is that we observed that the red color appeared in a specific popular clothing item (i.e. wedding dress) in winter. The red color with graphics is a typical wedding dress in China (cf. Fig. \ref{fig:eval_ex} (c)). This observation attracts our interests of embedding geographic information into the system in the future to further enable a more comprehensive framework. 

Furthermore, we are also interested in the changes in the clothing features trend. These changes in clothing features could indicate special clothing features for a specific season. The comparison is shown in Fig. \ref{fig:eval_trend}. In the style category, bags and accessories, belt, and placket increased substantially. In the color category, blue in the upper body and red in the upper and lower body showed an upward trend. In the style category, collar decreased markedly. In the color category, brown and gray in the lower body, and yellow in the upper body showed a downturn trend. An interesting observation is that there are more v-shape neckline in spring. The clothing products in Fig. \ref{fig:eval_ex} (a) are good demonstrations. Another interesting observation is that there are a popular short skirt and a popular coat in winter (cf. Fig. \ref{fig:eval_wrong} (b)). A reasonable explanation is that people tend to wear layers to make quick adjustments based on different indoor and outdoor environments. These particular clothing outfits could be discovered through our proposed framework. In summary, it seems clear that mining of selling clothing features in an online shopping website indeed has benefit to a big picture of clothing element preference, which could influence both clothing production and clothing consumption in a timely fashion. Furthermore, through our experimental results and observations, social conditions and natural conditions, as well as weather and culture, could be important factors for people to determine what they would be likely to wear and purchase. 

\section{Conclusions}
In this work, we organize and exploit a large-scale online shopping dataset in order to investigate the possible popular and attractive clothing features. In addition, we have developed machine learning based methods to automatically prune noisy images and detect clothing features as the representation of popular clothing style features. We conduct our experiments on two datasets and further demonstrate that the proposed framework is effective in discriminative mining of best-selling clothing features. In the future, we plan to integrate more clothing information (e.g. price or customer profile) and more clothing related datasets \cite{RW:liu2016deepfashion} to increase the comprehensive views of estimating popularity of clothing product for more comprehensive studies. Moreover, there is also a keen interest in exploring the proper clothing outfits to recommend users by aggregating user preference \cite{RW:he2016ups} in clothing style. One future direction would be to incorporate these features into the existing model and the scope can also be extended to emerging applications such as clothing advertising.

\section{Acknowledgments}
We gratefully acknowledge the Taiwan Government MOST Study Abroad Program grants 105-2917-I-564-060 and the support of New York State through the Goergen Institute for Data Science. 

\bibliographystyle{abbrv}
\bibliography{sigproc} 

\end{document}